# Dynamically Expanding Capacity of Autonomous Driving with Near-Miss Focused Training Framework

Ziyuan Yang[1*], Zhaoyang Li[2*], Jianming Hu[3†], and Yi Zhang[4]

[1]Dept. of Automation, Tsinghua Univ. Email: yangziyu22@mails.tsinghua.edu.cn
[2]Dept. of Automation, Tsinghua Univ. Email: lizhaoya21@mails.tsinghua.edu.cn
[3]Associate Professor, Dept. of Automation, Tsinghua Univ. Email: hujm@mail.tsinghua.edu.cn
[4]Professor, Dept. of Automation, Tsinghua Univ. Email: zhyi@tsinghua.edu.cn
[*]Work done with equal contribution.
[†]Corresponding author.

## ABSTRACT

The long-tail distribution of real driving data poses challenges for training and testing autonomous vehicles (AV), where rare yet crucial safety-critical scenarios are infrequent. And virtual simulation offers a low-cost and efficient solution. This paper proposes a near-miss focused training framework for AV. Utilizing the driving scenario information provided by sensors in the simulator, we design novel reward functions, which enable background vehicles (BV) to generate near-miss scenarios and ensure gradients exist not only in collision-free scenes but also in collision scenarios. And then leveraging the Robust Adversarial Reinforcement Learning (RARL) framework for simultaneous training of AV and BV to gradually enhance AV and BV capabilities, as well as generating near-miss scenarios tailored to different levels of AV capabilities. Results from three testing strategies indicate that the proposed method generates scenarios closer to near-miss, thus enhancing the capabilities of both AVs and BVs throughout training.

## INTRODUCTION

Autonomous driving technology has got rapid development these years. However, deploying it in the real world for widespread commercial use still poses safety challenges. Testing the safety of autonomous driving systems in real-world scenarios would require covering billions of miles (Feng et al. 2023). This is attributed to the phenomenon of a "*long-tail distribution*" in natural driving datasets, where the occurrence probability of useful safety-critical scenarios for testing autonomous driving systems is typically low. A primary approach to addressing this issue is through simulation platforms, artificially generating safety-critical scenarios, thereby expediting the training and testing of autonomous driving systems. However, in most current work, the Background Vehicle (BV) acting as the aggressor is aimed at generating collisions with the Autonomous Vehicle (AV), which serves as the subsequent subject for testing and training. Within the process of scenario generation, AV remains a fixed agent. This approach limits the capabilities of the trained BV as it cannot adapt its attack performance adaptively based on different abilities exhibited by AV. Moreover, the generated scenarios focus solely on generating collisions, with the gradient of danger existing primarily in factors such as the distance between BV and AV or their relative velocities, limited to the moments just before an actual collision occurs. However, collision scenarios also vary in severity. For instance, if an AV is surrounded by four BVs approaching from different directions simultaneously, the AV would need to practically take off to escape. Therefore, there should be a distinction in the severity of collision scenarios to measure the level of danger in each specific scenario.

Considering that the AV operates as a complicated learner, we propose integrating relevant concepts from educational psychology to aid in formulating training strategies. In educational psychology, there exists a concept known as the "Zone of Proximal Development" (ZPD), introduced by the Soviet psychologist Lev Vygotsky. It refers to the gap between a learner's current abilities and their potential abilities with guidance. Learners require assistance to bridge this gap in their abilities. Thus, a plausible learning model involves collaborative learning between two learners: one slightly more capable than the other (Wertsch et al. 1984). Analogous to the field of autonomous driving, the range of safe driving scenarios that an AV can handle independently represents the Zone of Actual Development (ZAD). The critical scenarios where the AV and BV experience severe collisions stand as the boundary beyond which the capabilities of the AV lie. Meanwhile, the scenarios where the BV poses dangerous yet non-lethal attacks on the AV, allowing the AV a possibility of escape, represent the zone within the AV's proximal development range.

Hence, we designate BV and AV in our training model as learners with slightly differing capabilities. Leveraging the framework of Robust Adversarial Reinforcement Learning (RARL) (Pinto et al. 2017), we devise a training framework where the AV and BV undergo alternating training sessions. Throughout each iteration, BV is trained to attack AV, but with the imposition of rewards at each step to regulate the intensity of the attack. This compels the generation of near-miss scenarios, subsequently training the AV to evade the BV's attacks and proceed with normal driving. Additionally, constraints on speed and steering are implemented to ensure that both the AV and BV adopt behaviors that closely mimic real-world driving situations.

Our primary contributions include: 1. Proposing a quantification method for evaluating the criticality of near-miss scenarios, thereby delineating the severity of critical situations. 2. Introducing a dynamic training approach for both the AV and BV, aiming to consistently expand the boundaries of their respective capabilities while maintaining stability.

## RELATED WORK

In the field of autonomous driving scenario generation, adversarial scenario generation stands out as a widely applied method. BV acting as the controlled agent, launch adversarial attacks against AV, thereby creating safety-critical scenarios used in the training and safety testing of autonomous driving system. Modeling scenes as Markov Decision Processes (MDP) and solving them through reinforcement learning (RL) represents a more commonly adopted approach. Feng et al. (2021) and Sun et al. (2021) utilizes Deep Q-Networks to generate discrete adversarial traffic scenarios. Kuutti et al. (2020) employs the Advantage Actor Critic (A2C) algorithm to control a BV in a following-car scenario. Niu et al. (2021) reinforces AV using adversarial models within deep reinforcement learning, gradually exposing them to increasingly challenging scenarios. (Re)2H2O proposed by Niu et al. (2023) employs a hybrid offline-online RL to generate safety-critical scenarios, enhancing scenario authenticity by integrating real-world data. Rempe et al. (2022) introduces the STRIVE framework, optimizing gradients to generate critical scenarios, increasing scenario diversity through adversarial learning methods.

Most existing scenario generation work aims to create collision scenarios, where the gradients of the scenarios exist primarily in non-collision scenarios. This is assessed through physical quantities like the minimum relative distance and relative velocity between two vehicles (Ding et al. 2020). Severity assessment for collision scenarios often centers on analyzing accident data, categorizing collision incidents based on the resulting property and personnel damage (Hungar et al. 2017), an aspect less explored in scenario generation efforts. However, in scenario generation, it's not that more dangerous scenarios are more beneficial, and AV needs to have a certain probability of escaping safety-critical scenarios in order to improve their capabilities. Hence, generating near-miss scenarios holds substantial value (Feng et al. 2023). Calò et al. (2020) design a search-based method to generate avoidable collision scenarios. Tuncali et al. (2019) propose a scenario generation method based on rapid random searches, guiding near-miss scenario generation through cost function design. In these approaches, AV is a fixed agent, limiting the transferability of generated scenarios to accommodate AV with different skill levels.

In order to dynamically generate corresponding near-miss scenarios during the process of enhancing AV capabilities, we propose a novel framework in this paper focusing on the generation of near-miss scenarios for autonomous driving training. Leveraging RARL, our framework dynamically trains AV and BVs. We measure the criticality of scenarios using collision momentum and obstacle frames, guiding BVs to generate near-miss scenarios. Through experiments, we demonstrate that our approach trains AV to drive more safely. Moreover, BVs can generate near-miss scenarios for AV of specific skill levels, assisting in the steady improvement of AV capabilities.

## METHODOLOGY

**Problem Formulation**
*a) Scenario Formulation*

Our scenarios take place within a specified length of a straight highway section. The traffic participants within the scenarios encompass one AV ($V_0$) and $N$ BVs ($V_i, i = 1, ..., N$) traveling on this road segment. For a specific moment $t$, the scene at the moment includes the state and action of all vehicles involved:

$$s_t = [s_t^0, s_t^1, ..., s_t^N]^T, \ a_t = [a_t^0, a_t^1, ..., a_t^N]^T$$

For each vehicle at moment $t$, the state vector comprises its lateral and longitudinal position coordinates $(x, y)$, velocity $v$, heading angle $\theta$, acceleration $\alpha$, and angular velocity $\omega$. The action vector consists of brake/throttle control parameter $p$ and steering wheel control parameter $\delta$:

$$s_t^i = [x_t^i, y_t^i, v_t^i, \theta_t^i, \alpha_t^i, \omega_t^i], \ a_t^i = [p_t^i, \delta_t^i]$$

According to the definitions in (Feng et al., 2020), we depict a scenario as a finite sequence of $H$-frame scenes:

$$s_0 \rightarrow u_0 \rightarrow s_1 \rightarrow u_0 \rightarrow \cdots \rightarrow s_H \rightarrow u_H$$

To simulate the dynamic state transitions in driving behavior as accurately as possible, we utilize the CARLA simulator (Dosovitskiy et al. 2017). We employ the kinematic model within CARLA to facilitate state transitions, denoted as $K$, to mimic dynamics of the real world:

$$s_{t+1} = K(s_t, a_t)$$

*b) Driving Policy Optimization*

In the problem formulation of RL, traffic scenarios can also be described using an MDP, often represented as a sextuple $(S, A, R, P, \rho, \gamma)$, where $S$ and $A$ are state space and action space that have been outlined above. $P, \rho$ and $\gamma$ are the state transition probability, initial state distribution and discount factor. The reward function at a given moment $t$ is defined as $r_t = r(s_t, a_t)$. Due to different objectives in policy optimization, the reward function is categorized into two types: $r_{drive}$ and $r_{attack}$. These correspond to training objectives for safe driving and adversarial attack, respectively. $r_{drive}$ includes the following items:

$$r_{drive} = \sum r_{collision, speed, yaw, cooperation}$$

- *Collision*: Prevent collisions between vehicles:

$$r_{collision} = -\alpha_1 \mathbf{I}_{\{if \ collision \ happens\}}$$

- *Speed*: Force speed fluctuates above and below the average speed:

$$r_{speed} = -\alpha_2 |v - \bar{v}|$$

where $\bar{v}$ represents the preset target average velocity.

- *Heading Direction*: Instruct vehicle to go straight:

$$r_{yaw} = -\alpha_3 |\theta + 90|$$

where heading angle for straight driving is -90 degrees.
- *Cooperation*: Encourage driving states among multiple vehicles to have minimal differences:

$$r_{cooperation} = -\alpha_4(v_{max} - v_{min})$$

Due to different training driving objectives, the composition of items in $r_{attack}$ and $r_{drive}$ differs:

$$r_{attack} = \sum r_{speed, distance, lane, obstacle, impulse}$$

- *Speed*: Encourage speed fluctuates above and below the average speed and drive more smoothly:

$$r_{speed} = -\alpha_5 |v_{adv} - \bar{v}|$$

where $\bar{v}$ represents the preset target average velocity.

- *Distance*: Encourage BVs to approach AV during driving:

$$r_{distance} = -\alpha_6 d_{min}$$

where $d_{min}$ represents the minimum distance between multiple BVs and AV.

- *Lane*: Force vehicles to maintain lane travel:

$$r_{lane} = -\alpha_7 \omega_{adv}$$

where $\omega_{adv}$ is the angular velocity of BV.

- *Obstacle*: Encourage BV to occlude AV for a longer duration of frames：

$$r_{obstacle} = \alpha_8 f + \alpha_8 (1 - \frac{d}{5}) \mathbf{I}_{\{if\ d < d_{min}\}}$$

where $f$ represents the number of frames in each step during which BV occludes AV, and $d$ is the current minimum distance between BV and AV in this step. If this distance is smaller than the historical minimum distance $d_{min}$, an additional reward is provided.

- *Impulse*: Instruct BVs to avoid collisions with AV that are too forceful：

$$r_{impulse} = -\alpha_9 J_{max}$$

where $J_{max}$ represents the maximum impulse of the collision between BVs and AV in current step.

We choose the offline RL algorithm Soft Actor-Critic (SAC) (Haarnoja et al. 2018) for iterating the driving strategy. In conventional Actor-Critic framework, Q-function approximation $\hat{Q}$ is obtained by minimizing the standard Bellman error. Meanwhile, the policy $\hat{\pi}$ is improved by maximizing the Q-function:

$$\hat{Q} \leftarrow \arg\min_Q \mathbb{E}_{\mathbf{s,a,s'} \sim \mathcal{U}} \left[ \frac{1}{2} \left( \left( Q - \hat{B}^\pi \hat{Q} \right)(\mathbf{s,a}) \right)^2 \right]$$

$$\hat{\pi} \leftarrow \arg\max_\pi \mathbb{E}_{\mathbf{s,a} \sim \mathcal{U}} \left[ \hat{Q}(\mathbf{s,a}) \right]$$

where $\mathcal{U}$ represents the data buffer produced by an earlier iteration of policy $\hat{\pi}$ via simulated interactions online. The Bellman evaluation operator $\widehat{\mathcal{B}}^\pi$ is defined as $\widehat{\mathcal{B}}^\pi \hat{Q}(\mathbf{s,a}) = r(\mathbf{s,a}) + \gamma \mathbb{E}_{\mathbf{a'} \sim \hat{\pi}(\mathbf{a'}|\mathbf{s'})} \hat{Q}(\mathbf{s',a'})$. While the objective function of SAC

includes an entropy term of the policy distribution:

$$J_\pi(\phi) = \sum_{t=0}^{H} \mathbb{E}_{s_0 \in \rho, a_t \sim \pi_\phi(\cdot|s_t), s_{t+1} \sim P(\cdot|s_t, a_t)} \left[ \gamma^t r(s_t, a_t) + \alpha \mathrm{H}\left(\pi_f(\cdot|s_t)\right) \right]$$

In this context, $\pi_\phi$ signifies the driving policy defined by $\phi$, with $\mathcal{H}$ denoting the entropy of distribution. The parameter $\alpha$ serves as the temperature hyperparameter within the algorithm. This setup facilitates exploring a broader range of potential actions while maximizing the expected reward discounted by $\gamma$, which boosts the model's resilience and aligns better with real-world traffic scenario applications.

To represent randomness, we depict the driving policy as a Gaussian distribution $\mathcal{N}(\mu, \sigma)$, where $\mu$ and $\sigma$ are approximated by a neural network $\mathcal{W}_\theta$ (Niu et al. 2023). Due to the constraints in CARLA, where throttle/brake and steering control inputs are limited between -1 and 1, we incorporate tanh and linear mapping functions to fit the driving policy.

**Algorithmic Framework**

Borrowing the idea of RARL (Pinto et al. 2017), we divide the training of AV and BVs into two stages. Initially, the policy parameters for both are sampled from random distributions.

Table 1 shows the process of algorithmic framework. In each of the $N_{iter}$ iterations, we alternately execute a two-step optimization process. Firstly, for $N_\mu$ iterations, the AV policy parameters remain unchanged as $\theta^v$, while the parameters $\theta^\mu$ of BVs' policy are optimized to maximize $r_{attack}$. For the next step, BV policy parameters $\theta^\mu$ are held constant for the next $N_v$ iterations. Before each parameter update, the *roll* function samples $N_{step}$ step samples under the current policy functions of AV and BV at that time, where $\varepsilon$ represents the driving environment in the simulator. This alternating process repeats for $N_{iter}$ iterations.

**Tabel 1: Pseudo code of the proposed algorithm**

---
**Algorithm 1** Algorithmic Framework
---
**Input:** Driving environment $\varepsilon$; BV policy $\mu$ and AV policy $v$
**Initialize:** Learnable parameters $\theta_0^\mu$ for $\mu$ and $\theta_0^v$ for $v$
**for** $i = 1, 2, \ldots, N_{iter}$ **do**

$\quad \theta_i^\mu \leftarrow \theta_{i-1}^\mu$

$\quad$**for** $j = 1, 2, \ldots, N_\mu$ **do**

$\quad\quad \left\{\left(s_t^i, a_t^{1i}, a_t^{2i}, r_t^{1i}, r_t^{2i}\right)\right\} \leftarrow roll\left(\varepsilon, \mu_{\theta_i^\mu}, v_{\theta_{i-1}^\mu}, N_{step}\right)$

$\quad\quad \theta_i^\mu \leftarrow policyOptimizer\left(\left\{\left(s_t^i, a_t^{1i}, r_t^{1i}\right), \mu, \theta_i^\mu\right\}\right)$

$\quad$**end for**
---

$$\theta_i^v \leftarrow \theta_{i-1}^v$$

    **for** $j = 1, 2, \ldots, N_v$ **do**

$$\{(s_t^i, a_t^{1i}, a_t^{2i}, r_t^{1i}, r_t^{2i})\} \leftarrow roll\left(\varepsilon, \mu_{\theta_i^\mu}, v_{\theta_i^\mu}, N_{step}\right)$$

$$\theta_i^v \leftarrow policyOptimizer\left(\{(s_t^i, a_t^{2i}, r_t^{2i}), \mu, \theta_i^v\}\right)$$

    **end for**

**end for**

**Return:** $\theta_{N_{iter}}^\mu$, $\theta_{N_{iter}}^v$

---

The diagram in Figure 1 provides a more intuitive representation of the training process. In the illustration, the grey-colored vehicles represent BVs, while the blue-colored one represent AV. Yellow markings around the vehicles indicate ongoing training, and the heptagonal stars represent the training aggressiveness strategy. The process flows from left to right. The first AV on the left has already undergone pretraining and convergence. Based on this, BVs are trained to learn normal driving strategies. Subsequently, the training alternates between BV aggressive strategies and AV normal driving strategies.

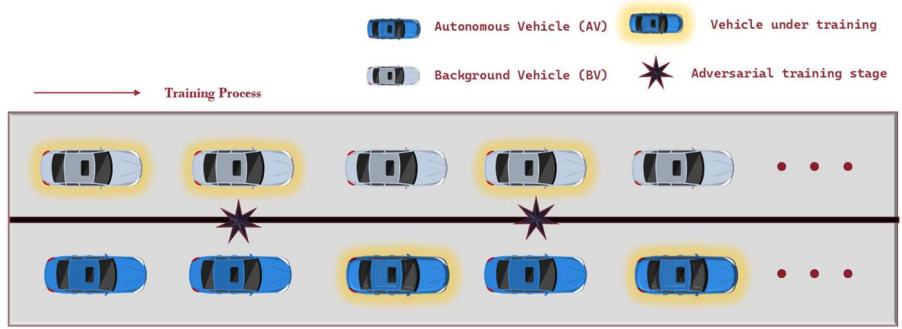

Figure 1. Training process schematic diagram.

## EXPERIMENTS

### Experiment Settings

*a) Environment Setup*

Due to the various sensors available in CARLA, providing diverse types of information for driving scenarios, and the closer resemblance of vehicle types and behavioral patterns to real-world driving, we choose CARLA as our simulation environment.

We select a straight segment of road in the "*Town04*" map of CARLA as our experimental area. Within 20 different starting point combinations, we randomly generated $n$ vehicles (1 AV, $n-1$ BVs) to drive in CARLA's *autopilot* mode to the departure area, which are considered as the initial state, ensuring both a reasonable and randomized starting point for the experiment. As for experimental parameters, according to experimental requirements and experience, we assign $\alpha_{1\sim9}$ values in reward function sequentially as [40, 0.5, 1.2, 0.5, 0.4, 1.5, 3, 10, 0.01].

*b) Evaluation Metrics*

To evaluate the driving performance of AV and BV, we select several evaluation metrics. To measure the temporal and spatial density of AV-BV collisions, we employ the metrics of **Average Collison Per Second (CPS)** and **Average Collision Frequency Per 100 Meter (CPM)** (Niu et al. 2023). These two metrics signify the average number of AV-BV collisions $N_{col}$ over the total testing time $T_{tt}$ and the total driving distance $D_{tt}$ of AV, respectively: $CPS = N_{col}/T_{tt}$, $CPM = N_{col}/D_{tt}$. Moreover, most existing work primarily design gradients in scenarios where collisions don't occur. To quantify gradients in collision scenarios, we use the **maximum impulse $J_{max}$** in collision scenarios to gauge the severity of the scenarios. And we use **obstacle frames (OBF)** in which BVs occlude AV to describe the extent of BVs' impact on AV's driving, thus measuring the criticality of scenarios from the perspective of AV driving, rather than simply based on collision occurrence.

*c) Testing Policies*

To test the policy models of AV and BVs, we select various testing strategies. The initial model for AV was a pretrained AV model (**RL_AV**), while for BV, we considered the BV model from CARLA autopilot mode (**auto_BV**), a pretrained BV model (**RL_BV**), and a BV model trained for 15 rounds with our proposed method (**trained_BV**). Thus we conduct three sets of experiments: 1) RL_AV vs auto_BV; 2) RL_AV vs RL_BV (double RL); 3) RL_AV vs trained_BV

**Experimental Results**

*a) Overall Experiments*

In this segment, we analyze the maximum impulse of collisions and OBF under different testing strategies. All RL AV and RL BV models undergo training and convergence before being configured to correspond to their respective training strategies. Figure 2 illustrates the comparison among three testing strategies and the proposed method when there are 3 BVs, and all results are computed using exponential smoothing with a 1.00 coefficient and averaged over 3 random seeds.

The maximum impulse reflects the intensity of collisions caused by BV to AV—larger impulses denote more severe collisions, while smaller impulses represent milder disturbances, indicating more closer to near-miss scenarios. OBF measures the degree of interference of BV to AV driving by reflecting the number of frames in which BV obstructs AV's view. A higher OBF suggests more obstruction to AV's normal driving by BV during operation.

In Figure 2(a), the proposed method during training maintains consistently low values for maximum impulse, indicating relatively mild collision impacts. RL_AV vs auto_BV, due to BV being set in autopilot mode, actively avoids collisions with AV, hence maintaining lowest values for maximum impulse. In the double RL setting, where both AV and BV are RL models, the collision frequency is higher due to the less efficient simultaneous training of multiple RL models, resulting in consistently higher impulse values. For the BV model trained with the proposed method for 10 rounds, which means $N_{iter}$ =10 in Table 1, its capabilities are enhanced, causing initially higher impulse values when paired with the AV pretrained model. However, over the course of training, these values gradually decrease and converge to a relatively high but stable level.

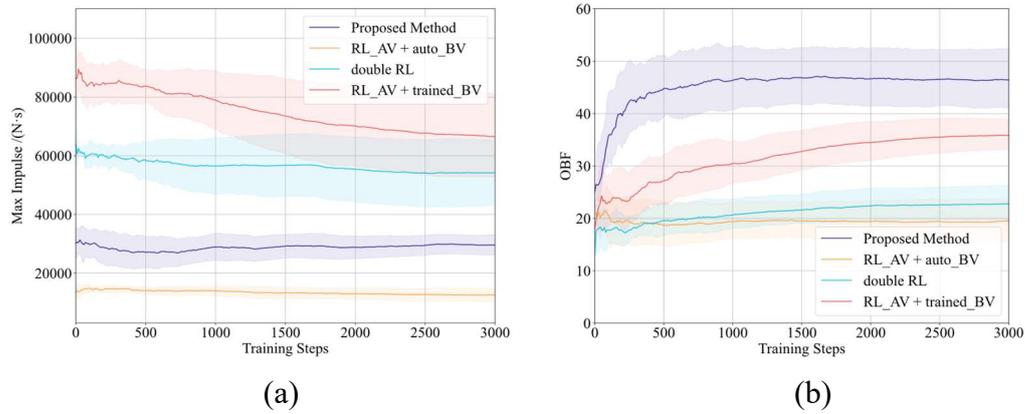

(a)            (b)

**Figure 2. Comparison between four training strategies.**

In Figure 2(b), the BV trained using the proposed method demonstrates a notable capability in creating substantial occlusion interference to AV, maintaining a comparatively high OBF after an initial rise. When BV operates in autopilot mode, resulting in no aggressive behavior towards AV, the OBF remains relatively low. The double RL model tends to end scenarios early due to collisions occurring at an earlier stage, resulting in lower OBF. When BV is the trained model, initially, due to its relatively aggressive nature towards AV, collisions occur earlier, leading to a lower OBF. As training steps increase and AV's capabilities improve, scenario durations increase, subsequently elevating the OBF, eventually converging to a relatively high level.

b) *Controlled Experiments*

In this section, we conduct tests using BV trained over different rounds (Round = 1, 2, ..., 15) along with a pretrained AV and AV trained over different rounds along with a BV after training 15 rounds. Figure 3 illustrates the changes in CPS (Collisions per Scene) and CPM (Collisions per Minute) as BV or AV training rounds increase.

From Figure 3(a), it is evident that as the BV training rounds increase, its aggressiveness towards pretrained AV also escalates. Both CPS and CPM show an upward trend, indicating an increase in the frequency of collisions per second and per hundred meters. And from Figure 3(b), when pretrained AV encounters trained

15 rounds BV, it is obvious that CPS and CPM decrease as AV training rounds increase, which means that BV will cause more collisions to AV at lower training procedure.

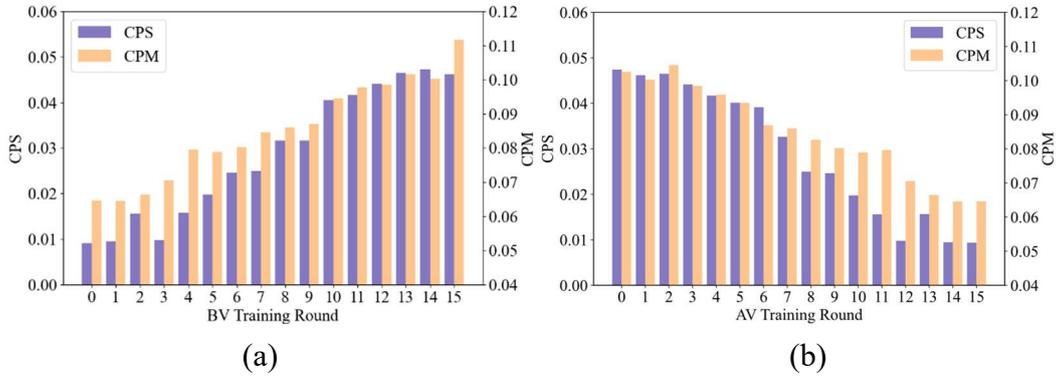

(a)                                                 (b)

**Figure 3. Controlled experiment results of varying-capability AV and BV**

*c) Scenarios in Simulation*

In this section, we present some scenarios during training rendered in the CARLA simulator to better illustrate the training outcomes. The blue one in Figure 4 is AV, and grey ones are BVs.

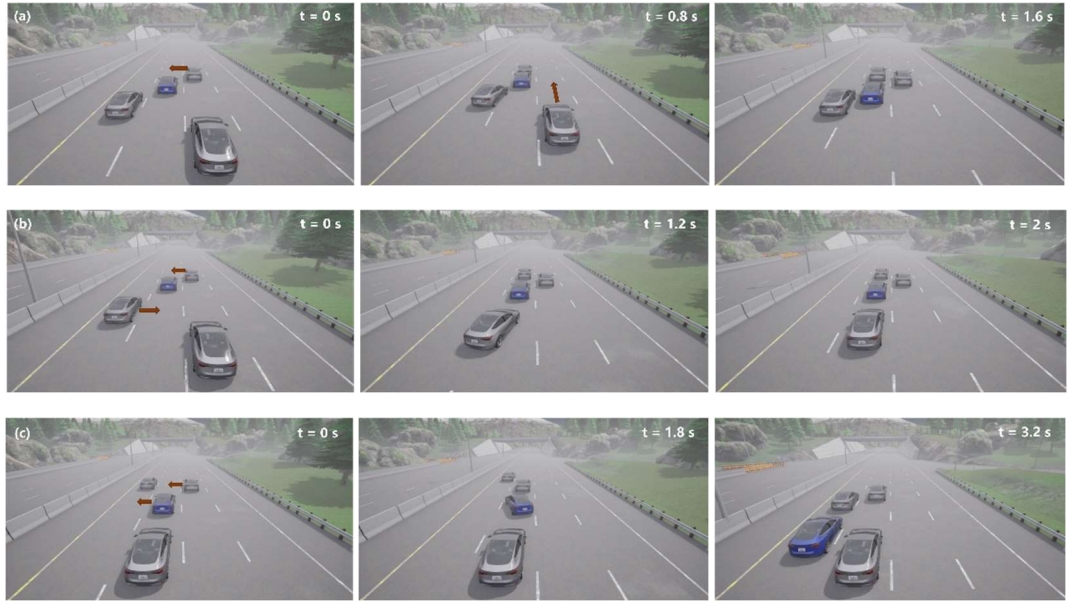

**Figure 4. Typical scenarios in simulation**

Figure 4(a) illustrates the BV in the right front of AV changing lanes, obstructing AV. Subsequently, the BV on the left rear side of AV accelerates, slightly turns towards AV, closely follows it, interfering with its lane-changing to evade the front BV. In Figure 4(b), both the BV on the right front and left rear of AV change lanes simultaneously. Meanwhile, the BV on the right rear accelerates, blocking AV from three directions. Figure 4(c) portrays a scenario where a vehicle in the right front of AV changes lanes. AV attempts to swerve to the left, but then a BV behind AV's original lane accelerates, obstructing AV, preventing it from returning to its initial lane.

**CONCLUSION**

This paper introduces a near-miss focused autonomous driving training framework. By designing reinforcement learning rewards, it establishes the foundation for generating near-miss scenarios, ensuring gradients exist not only in collision-free scenes but also in collision scenarios. Leveraging the RARL framework for training both AV and BV, it consistently enhances AV capabilities while enabling BV to purposefully generate near-miss scenarios to aid AV training. Experimental results demonstrate the effectiveness of this method in significantly improving AV capabilities. Moreover, BV's aggressiveness gradually increases during training. Compared to other training strategies, the proposed method generates scenarios that closely resemble near-miss situations. Future research could consider incorporating a wider array of traffic participants to enhance the similarity between simulated scenarios and real-world settings.

**ACKNOWLEDGEMENTS**

This work is supported by National Natural Science Foundation of China under Grant No. 62333015 and Beijing Natural Science Foundation L231014.